# Kernel-based Conditional Independence Test and Application in Causal Discovery

**Kun Zhang**  **Jonas Peters**  **Dominik Janzing**  **Bernhard Schölkopf**

Max Planck Institute for Intelligent Systems

Spemannstr. 38, 72076 Tübingen

Germany

## Abstract

Conditional independence testing is an important problem, especially in Bayesian network learning and causal discovery. Due to the curse of dimensionality, testing for conditional independence of continuous variables is particularly challenging. We propose a Kernel-based Conditional Independence test (KCI-test), by constructing an appropriate test statistic and deriving its asymptotic distribution under the null hypothesis of conditional independence. The proposed method is computationally efficient and easy to implement. Experimental results show that it outperforms other methods, especially when the conditioning set is large or the sample size is not very large, in which case other methods encounter difficulties.

## 1 Introduction

Statistical independence and conditional independence (CI) are important concepts in statistics, artificial intelligence, and related fields (Dawid, 1979). Let $X$, $Y$ and $Z$ denote sets of random variables. The CI between $X$ and $Y$ given $Z$, denoted by $X \perp\!\!\!\perp Y | Z$, reflects the fact that given the values of $Z$, further knowing the values of $X$ (or $Y$) does not provide any additional information about $Y$ (or $X$). Independence and CI play a central role in causal discovery and Bayesian network learning (Pearl, 2000; Spirtes et al., 2001; Koller and Friedman, 2009). Generally speaking, the CI relationship $X \perp\!\!\!\perp Y | Z$ allows us to drop $Y$ when constructing a probabilistic model for $X$ with $(Y, Z)$, which results in a parsimonious representation.

Testing for CI is much more difficult than that for unconditional independence (Bergsma, 2004). For CI tests, traditional methods either focus on the discrete case, or impose simplifying assumptions to deal with the continuous case – in particular, the variables are often assumed to have linear relations with additive Gaussian errors. In that case, $X \perp\!\!\!\perp Y | Z$ reduces to zero partial correlation or zero conditional correlation between $X$ and $Y$ given $Z$, which can be easily tested (for the links between partial correlation, conditional correlation, and CI, see Lawrance (1976)). However, nonlinearity and non-Gaussian noise are frequently encountered in practice, and hence this assumption can lead to incorrect conclusions.

Recently, practical methods have been proposed for testing CI for continuous variables without assuming a functional form between the variables as well as the data distributions, which is the case we are concerned with in this paper. To our knowledge, the existing methods fall into four categories. The first category is based on explicit estimation of the conditional densities or their variants. For example, Su and White (2008) define the test statistic as some distance between the estimated conditional densities $p(X|Y, Z)$ and $p(X|Y)$, and Su and White (2007) exploit the difference between the characteristic functions of these conditional densities. The estimation of the conditional densities or related quantities is difficult, which deteriorates the testing performance especially when the conditioning set $Z$ is not small enough. Methods in the second category, such as Margaritis (2005) and Huang (2010), discretize the conditioning set $Z$ to a set of bins, and transform CI to the unconditional one in each bin. Inevitably, due to the curse of dimensionality, as the conditioning set becomes larger, the required sample size increases dramatically. Methods in the third category, including Linton and Gozalo (1997) and Song (2009), provide slightly weaker tests than that for CI. For instance, the method proposed by Song (2009) tests whether one can find some (nonlinear) function $h$ and parameters $\theta_0$, such that $X$ and $Y$ are conditionally independent given a single index function $\lambda_{\theta_0}(Z) = h(Z^T \theta_0)$ of $Z$. In general, this is different from the test for $X \perp\!\!\!\perp Y | Z$: to see this, consider the case where $X$ and $Y$ depend on two different but overlapping subsets of $Z$; even if $X \perp\!\!\!\perp Y|Z$, it is impossible to find $\lambda_{\theta_0}(Z)$ given which $X$ and $Y$ are conditionally independent.

Fukumizu et al. (2004) give a general nonparametric characterization of CI using covariance operators in the reproducing kernel Hilbert spaces (RKHS), which inspired a kernel-based measure of conditional dependence (see also Fukumizu et al. (2008)). However, the distribution of this measure under the CI hypothesis is unknown, and consequently, it could not directly serve as a CI test. To get a test, one has to combine this conditional dependence measure with local bootstrap, or local permutation, which is used to determine the rejector region (Fukumizu et al., 2008; Tillman et al., 2009). This leads to the method in the fourth category. We denote it by $CI_{PERM}$. Like the methods in the second category, this approach would require a large sample size and tends to be unreliable when the number of conditioning variables increases.

In this paper we aim to develop a CI testing method which avoids the above drawbacks. In particular, based on appropriate characterizations of CI, we define a simple test statistic which can be easily calculated from the kernel matrices associated with $X$, $Y$, and $Z$, and we further derive its asymptotic distribution under the null hypothesis. We also provide ways to estimate such a distribution, and finally CI can be tested conveniently. This results in a Kernel-based Conditional Independence test (KCI-test). In this procedure we do not explicitly estimate the conditional or joint densities, nor discretize the conditioning variables. Our method is computationally appealing and is less sensitive to the dimensionality of $Z$ compared to other methods. Our results contain unconditional independence testing (similar to (Gretton et al., 2008)) as a special case.

## 2 Characterization of Independence and Conditional Independence

We introduce the following notational convention. Throughout this paper, $X$, $Y$, and $Z$ are continuous random variabels or sets of continuous random variables, with domains $\mathcal{X}$, $\mathcal{Y}$, and $\mathcal{Z}$, respectively. Define a measurable, positive definite kernel $k_\mathcal{X}$ on $\mathcal{X}$ and denote the corresponding RKHS by $\mathcal{H}_\mathcal{X}$. Similarly we define $k_\mathcal{Y}$, $\mathcal{H}_\mathcal{Y}$, $k_\mathcal{Z}$, and $\mathcal{H}_\mathcal{Z}$. In this paper we assume that all involved RKHS's are separable and square integrable. The probability law of $X$ is denoted by $P_X$, and similarly for the joint probability laws such as $P_{XZ}$. The spaces of square integrable functions of $X$ and $(X, Z)$ are denoted by $L_X^2$ and $L_{XZ}^2$, respectively. E.g., $L_{XZ}^2 = \{g(X,Z) \,|\, \mathbb{E}[g^2] < \infty\}$. $\mathbf{x} = \{x_1, ..., x_n\}$ denotes the i.i.d. sample of $X$ of size $n$. $\mathbf{K}_X$ is the kernel matrix of the sample $\mathbf{x}$, and the corresponding centralized kernel matrix is $\widetilde{\mathbf{K}}_X \triangleq \mathbf{H}\mathbf{K}_X\mathbf{H}$, where $\mathbf{H} = \mathbf{I} - \frac{1}{n}\mathbf{1}\mathbf{1}^T$ with $\mathbf{I}$ and $\mathbf{1}$ being the $n \times n$ identity matrix and the vector of 1's, respectively. By default we use the Gaussian kernel, i.e., the $(i,j)$th entry of $\mathbf{K}_X$ is $k(\mathbf{x}_i, \mathbf{x}_j) = \exp(-\frac{||\mathbf{x}_i - \mathbf{x}_j||^2}{2\sigma_X^2})$, where $\sigma_X$ is the kernel width. Similar notations are used for $Y$ and $Z$.

The problem we consider here is to test for CI between sets of continuous variables $X$ and $Y$ given $Z$ from their observed i.i.d. samples, without making cspecific assumptions on their distributions or the functional forms between them. $X$ and $Y$ are said to be conditionally independent given $Z$ if and only if $p_{X|Y,Z} = p_{X|Z}$ (or equivalently, $p_{Y|X,Z} = p_{Y|Z}$, or $p_{XY|Z} = p_{X|Z}p_{Y|Z}$). Therefore, a direct way to assess if $X \perp\!\!\!\perp Y|Z$ is to estimate certain probability densities and then evaluate if the above equation is plausible. However, density estimation in high dimensions is a difficult problem: it is well known that in nonparametric joint or conditional density estimation, due to the curse of dimensionality, to achieve the same accuracy the number of required data points is exponentially increasing in the data dimension. Fortunately, conditional (in)dependence is just one particular property associated with the distributions; to test for it, it is possible to avoid explicitly estimating the densities.

There are other ways to characterize the CI relation that do not explicitly involve the densities or the variants, and they may result in more efficient methods for CI testing. Recently, a characterization of CI is given in terms of the cross-covariance operator $\Sigma_{YX}$ on RKHS (Fukumizu et al., 2004). For the random vector $(X,Y)$ on $\mathcal{X} \times \mathcal{Y}$, the cross-covariance operator from $\mathcal{H}_\mathcal{X}$ to $\mathcal{H}_\mathcal{Y}$ is defined by the relation:

$$\langle g, \Sigma_{YX} f \rangle = \mathbb{E}_{XY}[f(X)g(Y)] - \mathbb{E}_X[f(X)]\mathbb{E}_Y[g(Y)]$$

for all $f \in \mathcal{H}_\mathcal{X}$ and $g \in \mathcal{H}_\mathcal{Y}$. The conditional cross-covariance operator of $(X,Y)$ given $Z$ is further defined by [1]

$$\Sigma_{YX|Z} = \Sigma_{YX} - \Sigma_{YZ}\Sigma_{ZZ}^{-1}\Sigma_{ZX}. \quad (1)$$

Intuitively, one can interpret it as the partial covariance between $\{f(X), \forall f \in \mathcal{H}_\mathcal{X}\}$ and $\{g(Y), \forall g \in \mathcal{H}_\mathcal{Y}\}$ given $\{h(Z), \forall h \in \mathcal{H}_\mathcal{Z}\}$.

If characteristic kernels[2] are used, the conditional

---
[1] If $\Sigma_{ZZ}$ is not invertible, one should use the right inverse instead of the inverse; see Corollary 3 in Fukumizu et al. (2004).

[2] A kernel $k_\mathcal{X}$ is said to be *characteristic* if the condition $E_{X\sim P}[f(X)] = E_{X\sim Q}[f(X)]$ ($\forall f \in \mathcal{H}$) implies $P = Q$, where $P$ and $Q$ are two probability distributions of $X$ (Fukumizu et al., 2008). Hence, the notion "characteristic" was also called "probability-determining" in Fukumizu et al. (2004). Many popular kernels, such as the Gaussian one, are charateristic.

cross-covariance operator is related to the CI relation, as seen from the following lemma.

**Lemma 1 [Characterization based on conditional cross-covariance operators (Fukumizu et al., 2008)]**
Denote $\ddot{X} \triangleq (X, Z)$, $k_{\ddot{X}} \triangleq k_{\mathcal{X}} k_{\mathcal{Z}}$, and $\mathcal{H}_{\ddot{X}}$ the RKHS corresponding to $k_{\ddot{X}}$. Assume $\mathcal{H}_{\mathcal{X}} \subset L_X^2$, $\mathcal{H}_{\mathcal{Y}} \subset L_Y^2$, and $\mathcal{H}_{\mathcal{Z}} \subset L_Z^2$. Further assume that $k_{\ddot{X}} k_{\mathcal{Y}}$ is a characteristic kernel on $(\mathcal{X} \times \mathcal{Y}) \times \mathcal{Z}$, and that $\mathcal{H}_{\mathcal{Z}} + \mathbb{R}$ (the direct sum of the two RKHSs) is dense in $L^2(P_Z)$. Then
$$\Sigma_{\ddot{X}Y|Z} = 0 \iff X \perp\!\!\!\perp Y|Z. \quad (2)$$

Note that one can replace $\Sigma_{\ddot{X}Y|Z}$ with $\Sigma_{\ddot{X}\ddot{Y}|Z}$, where $\ddot{Y} \triangleq (Y, Z)$, in the above lemma. Alternatively, Daudin (1980) gives the characterization of CI by *explicitly enforcing the uncorrelatedness of functions in suitable spaces*, which may be intuitively more appealing. In particular, consider the constrained $L^2$ spaces

$$\begin{aligned}
\mathcal{E}_{XZ} &\triangleq \{\tilde{f} \in L_{XZ}^2 \mid \mathbb{E}(\tilde{f}|Z) = 0\}, \\
\mathcal{E}_{YZ} &\triangleq \{\tilde{g} \in L_{YZ}^2 \mid \mathbb{E}(\tilde{g}|Z) = 0\}, \\
\mathcal{E}'_{YZ} &\triangleq \{\tilde{g}' \mid \tilde{g}' = g'(Y) - \mathbb{E}(g'|Z),\ g' \in L_Y^2\}.(3)
\end{aligned}$$

They can be constructed from the corresponding $L^2$ spaces via nonlinear regression. For instance, for any function $f \in L_{XZ}^2$, the corresponding function $\tilde{f}$ is given by
$$\tilde{f}(\ddot{X}) = f(\ddot{X}) - \mathbb{E}(f|Z) = f(\ddot{X}) - h_f^*(Z), \quad (4)$$

where $h_f^*(Z) \in L_Z^2$ is the regression function of $f(\ddot{X})$ on $Z$. One can then relate CI to uncorrelatedness in the following way.

**Lemma 2 [Characterization based on partial association (Daudin, 1980)]**
*The following conditions are equivalent to each other.*
*(i.)* $X \perp\!\!\!\perp Y|Z$; *(ii.)* $\mathbb{E}(\tilde{f}\tilde{g}) = 0$, $\forall \tilde{f} \in \mathcal{E}_{XZ}$ and $\tilde{g} \in \mathcal{E}_{YZ}$; *(iii.)* $\mathbb{E}(\tilde{f}g) = 0$, $\forall \tilde{f} \in \mathcal{E}_{XZ}$ and $g \in L_{YZ}^2$; *(iv.)* $\mathbb{E}(\tilde{f}\tilde{g}') = 0$, $\forall \tilde{f} \in \mathcal{E}_{XZ}$ and $\tilde{g}' \in \mathcal{E}'_{YZ}$; *(v.)* $\mathbb{E}(\tilde{f}g') = 0$, $\forall \tilde{f} \in \mathcal{E}_{XZ}$ and $g' \in L_Y^2$.

The above result can be considered as a generalization of the partial correlation based characterization of CI for Gaussian variables. Suppose that $(X, Y, Z)$ is jointly Gaussian; then $X \perp\!\!\!\perp Y|Z$ is equivalent to the vanishing of the partial correlation coefficient $\rho_{XY \cdot Z}$. Here, intuitively speaking, condition *(ii)* means that any "residual" function of $(X, Z)$ given $Z$ is uncorrelated from that of $(Y, Z)$ given $Z$. Note that $\mathcal{E}_{XZ}$ (resp. $\mathcal{E}_{YZ}$) contains all functions of $X$ and $Z$ (resp. of $Y$ and $Z$) that cannot be "explained" by $Z$, in the sense that any function $\tilde{f} \in \mathcal{E}_{XZ}$ (resp. $\tilde{g} \in \mathcal{E}_{YZ}$) is uncorrelated with any function of $Z$ (Daudin, 1980).

From the definition of the conditional cross-covariance operator (1), one can see the close relationship between the conditions in Lemma 1 and those in Lemma 2. However, Lemma 1 has practical advantages: in Lemma 2 one has to consider all functions in $L^2$, while Lemma 1 exploits the spaces corresponding to some characteristic kernels, which might be much smaller. In fact, if we restrict the functions $f$ and $g'$ to the spaces $\mathcal{H}_{\ddot{X}}$ and $\mathcal{H}_{\mathcal{Y}}$, respectively, Lemma 2 is then reduced to Lemma 1. The above characterizations of CI motivated our statistic for testing $X \perp\!\!\!\perp Y|Z$, as presented below.

## 3 A Kernel-Based Conditional Independence Test

### 3.1 General results

As seen above, independence and CI can be characterized by uncorrelatedness between functions in certain spaces. We first give some general results on the asymptotic distributions of some statistics defined in terms of the kernel matrices under the condition of such uncorrelatedness. Later those results will be used for testing for CI as well as unconditional independence.

Suppose that we are given the i.i.d. samples $\mathbf{x} \triangleq (x_1, ..., x_t, ..., x_n)$ and $\mathbf{y} \triangleq (y_1, ..., y_t, ..., y_n)$ for $X$ and $Y$, respectively. Suppose further that we have the eigenvalue decompositions (EVD) of the centralized kernel matrices $\widetilde{\mathbf{K}}_X$ and $\widetilde{\mathbf{K}}_Y$, i.e., $\widetilde{\mathbf{K}}_X = \mathbf{V}_\mathbf{x} \mathbf{\Lambda}_\mathbf{x} \mathbf{V}_\mathbf{x}^T$ and $\widetilde{\mathbf{K}}_Y = \mathbf{V}_\mathbf{y} \mathbf{\Lambda}_\mathbf{y} \mathbf{V}_\mathbf{y}^T$, where $\mathbf{\Lambda}_\mathbf{x}$ and $\mathbf{\Lambda}_\mathbf{y}$ are the diagonal matrices containing the non-negative eigenvalues $\lambda_{\mathbf{x},i}$ and $\lambda_{\mathbf{y},i}$, respectively. Here, the eigenvalues are sorted in descending order, i.e., $\lambda_{\mathbf{x},1} \geq \lambda_{\mathbf{x},2} \geq \ldots \geq \lambda_{\mathbf{x},i} \geq 0$, and $\lambda_{\mathbf{y},1} \geq \lambda_{\mathbf{y},2} \geq \ldots \geq \lambda_{\mathbf{y},i} \geq 0$. Let $\boldsymbol{\psi}_\mathbf{x} = [\psi_1(\mathbf{x}), ..., \psi_n(\mathbf{x})] \triangleq \mathbf{V}_\mathbf{x} \mathbf{\Lambda}_\mathbf{x}^{1/2}$ and $\boldsymbol{\phi}_\mathbf{y} = [\phi_1(\mathbf{y}), ..., \phi_n(\mathbf{y})] \triangleq \mathbf{V}_\mathbf{y} \mathbf{\Lambda}_\mathbf{y}^{1/2}$. I.e., $\psi_i(\mathbf{x}) = \sqrt{\lambda_{\mathbf{x},i}} \mathbf{V}_{\mathbf{x},i}$, where $\mathbf{V}_{\mathbf{x},i}$ denotes the $i$th eigenvector of $\widetilde{\mathbf{K}}_X$.

On the other hand, consider the eigenvalues $\lambda_{X,i}^*$ and eigenfunctions $u_{X,i}$ of the kernel $k_{\mathcal{X}}$ w.r.t. the probability measure with the density $p(x)$, i.e., $\lambda_{X,i}^*$ and $u_{X,i}$ satisfy
$$\int k_{\mathcal{X}}(x, x') u_{X,i}(x) p(x) dx = \lambda_{X,i}^* u_{X,i}(x').$$

Here we assume that $u_{X,i}$ have unit variance, i.e., $\mathbb{E}[u_{X,i}^2(X)] = 1$. We also sort $\lambda_{X,i}^*$ in descending order. Similarly, we define $\lambda_{Y,i}^*$ and $u_{Y,i}$ of $k_{\mathcal{Y}}$. Define
$$S_{ij} \triangleq \frac{1}{\sqrt{n}} \psi_i(\mathbf{x})^T \phi_j(\mathbf{y}) = \frac{\sum_{t=1}^n \psi_i(x_t) \phi_j(y_t)}{\sqrt{n}}$$

with $\psi_i(x_t)$ being the $t$-th component of the vector $\psi_i(\mathbf{x})$. We then have the following results.

**Theorem 3** *Suppose that we are given arbitrary centred kernels $k_\mathcal{X}$ and $k_\mathcal{Y}$ with discrete eigenvalues and the corresponding RKHS's $\mathcal{H}_\mathcal{X}$ and $\mathcal{H}_\mathcal{Y}$ for sets of random variables $X$ and $Y$, repectively. We have the following three statements.*

*1) Under the condition that $f(X)$ and $g(Y)$ are uncorrelated for all $f \in \mathcal{H}_\mathcal{X}$ and $g \in \mathcal{H}_\mathcal{Y}$, for any $L$ such that $\lambda^*_{X,L+1} \neq \lambda^*_{X,L}$ and $\lambda^*_{Y,L+1} \neq \lambda^*_{Y,L}$, we have*

$$\sum_{i,j=1}^{L} S_{ij}^2 \xrightarrow{d} \sum_{k=1}^{L^2} \mathring{\lambda}^*_k z_k^2, \text{ as } n \to \infty, \quad (5)$$

*where $z_k$ are i.i.d. standard Gaussian variables (i.e., $z_k^2$ are i.i.d. $\chi_1^2$-distributed variables), $\mathring{\lambda}^*_k$ are the eigenvalues of $\mathbb{E}(\mathbf{w}\mathbf{w}^T)$ and $\mathbf{w}$ is the random vector obtained by stacking the $L \times L$ matrix $\mathbf{N}$ whose $(i,j)$th entry is $N_{ij} = \sqrt{\lambda^*_{X,i} \lambda^*_{Y,j}} u_{X,i}(X) u_{Y,j}(Y)$.[3]*

*2) In particular, if $X$ and $Y$ are further independent, we have*

$$\sum_{i,j=1}^{L} S_{ij}^2 \xrightarrow{d} \sum_{i,j=1}^{L} \lambda^*_{X,i} \lambda^*_{Y,j} \cdot z_{ij}^2, \text{ as } n \to \infty, \quad (6)$$

*where $z_{ij}^2$ are i.i.d. $\chi_1^2$-distributed variables.*

*3) The results (5) and (6) also hold if $L = n \to \infty$.*

All proofs are sketched in Appendix. We note that

$$\text{Tr}(\widetilde{\mathbf{K}}_X \widetilde{\mathbf{K}}_Y) = \text{Tr}(\boldsymbol{\psi}_\mathbf{x} \boldsymbol{\psi}_\mathbf{x}^T \boldsymbol{\psi}_\mathbf{y} \boldsymbol{\psi}_\mathbf{y}^T)$$
$$= \text{Tr}(\boldsymbol{\psi}_\mathbf{x}^T \boldsymbol{\psi}_\mathbf{y} \boldsymbol{\psi}_\mathbf{y}^T \boldsymbol{\psi}_\mathbf{x}) = n \sum_{i,j=1}^{n} S_{ij}^2. \quad (7)$$

Hence, the above theorem gives the asymptotic distribution of $\frac{1}{n}\text{Tr}(\widetilde{\mathbf{K}}_X \widetilde{\mathbf{K}}_Y)$ under the condition that $f(X)$ and $g(Y)$ are always uncorrelated. Combined with the characterizations of (conditional) independence, this inspires the corresponding testing method. We further give the following remarks on Theorem 3. First, $X$ and $Y$ are not necessarily disjoint, and $\mathcal{H}_\mathcal{X}$ and $\mathcal{H}_\mathcal{Y}$ can be any RKHS's. Second, in practice the eigenvalues $\mathring{\lambda}^*_k$ are not known, and one needs to use the empirical ones instead, as discussed below.

### 3.2 Unconditional independence testing

As a direct consequence of the above theorem, we have the following result which allows for kernel-based unconditional independence testing.

---
[3]Equivalently, one can consider $\mathring{\lambda}^*_k$ as the eigenvalues of the tensor $\mathcal{T}$ with $\mathcal{T}_{ijkl} = \mathbb{E}(N_{ij,t} N_{kl,t})$.

**Theorem 4 [Independence test]**
*Under the null hypothesis that $X$ and $Y$ are statistically independent, the statistic*

$$T_{UI} \triangleq \frac{1}{n} Tr(\widetilde{\mathbf{K}}_X \widetilde{\mathbf{K}}_Y) \quad (8)$$

*has the same asymptotic distribution as*

$$\check{T}_{UI} \triangleq \frac{1}{n^2} \sum_{i,j=1}^{n} \lambda_{\mathbf{x},i} \lambda_{\mathbf{y},j} z_{ij}^2, \quad (9)$$

*i.e., $T_{UI} \xrightarrow{d} \check{T}_{UI}$ as $n \to \infty$.*

This theorem inspires the following unconditional independence testing procedure. Given the samples $\mathbf{x}$ and $\mathbf{y}$, one first calculates the centralized kernel matrices $\widetilde{\mathbf{K}}_X$ and $\widetilde{\mathbf{K}}_Y$ and their eigenvalues $\lambda_{\mathbf{x},i}$ and $\lambda_{\mathbf{y},i}$, and then evaluates the statistic $T_{UI}$ according to (8). Next, the empirical null distribution of $\check{T}_{UI}$ under the null hypothesis can be simulated in the following way: one draws i.i.d. random samples from the $\chi_1^2$-distributed variables $z_{ij}^2$, and then generates samples for $\check{T}_{UI}$ according to (9). (Later we will give another way to approximate the asympotic null distribution.) Finally the $p$-value can be found by locating $T_{UI}$ in the null distribution.

This unconditional independence testing method is closely related to the one based on the Hilbert-Schmidt independence criterion (HSIC) proposed by Gretton et al. (2008). Actually the defined statistics (our statistic $T_{UI}$ and $HSIC_b$ in Gretton et al. (2008)) are the same, but the asymptotic distributions are in different forms. In our results the asymptotic distribution only involves the eigenvalues of the two regular kernel matrices $\widetilde{\mathbf{K}}_X$ and $\widetilde{\mathbf{K}}_Y$, while in their results it depends on the eigenvalues of an order-four tensor, which are more difficult to calculate.

### 3.3 Conditional independence testing

Here we would like to make use of condition $(iv)$ in Lemma 2 to test for CI, but the considered functional spaces are $f(\ddot{X}) \in \mathcal{H}_{\ddot{\mathcal{X}}}$, $g'(Y) \in \mathcal{H}_\mathcal{Y}$, $h^*_f(Z) \in \mathcal{H}_\mathcal{Z}$, and $h^*_{g'}(Z) \in \mathcal{H}_\mathcal{Z}$, as in Lemma 1. The functions $\tilde{f}(\ddot{X})$ and $\tilde{g}'(Y,Z)$ appearing in condition $(iv)$, whose spaces are denoted by $\mathcal{H}_{\ddot{\mathcal{X}}|\mathcal{Z}}$ and $\mathcal{H}_{\mathcal{Y}|\mathcal{Z}}$, respectively, can be constructed from the functions $f$, $g'$, $h^*_f$, and $h^*_{g'}$.

Suppose that we already have the centralized kernel matrices $\widetilde{\mathbf{K}}_{\ddot{X}}$ (which is the centralized kernel matrix of $\ddot{X} = (X,Z)$), $\widetilde{\mathbf{K}}_Y$, and $\widetilde{\mathbf{K}}_Z$ on the samples $\mathbf{x}$, $\mathbf{y}$, and $\mathbf{z}$. We use kernel ridge regression to estimate the regression function $h^*_f(Z)$ in (4), and one can easily see that $\hat{h}^*_f(\mathbf{z}) = \widetilde{\mathbf{K}}_Z (\widetilde{\mathbf{K}}_Z + \varepsilon \mathbf{I})^{-1} \cdot f(\ddot{\mathbf{x}})$, where $\varepsilon$ is a small positive regularization parameter (Schölkopf and Smola, 2002). Consequently, $\tilde{f}(\ddot{\mathbf{x}})$ can be constructed as $\tilde{f}(\ddot{\mathbf{x}}) = f(\ddot{\mathbf{x}}) - \hat{h}_f^*(\mathbf{z}) = \mathbf{R}_Z \cdot f(\ddot{\mathbf{x}})$, where

$$\mathbf{R}_z = \mathbf{I} - \widetilde{\mathbf{K}}_Z(\widetilde{\mathbf{K}}_Z + \varepsilon\mathbf{I})^{-1} = \varepsilon(\widetilde{\mathbf{K}}_Z + \varepsilon\mathbf{I})^{-1}. \quad (10)$$

Based on the EVD decoposition $\widetilde{\mathbf{K}}_{\ddot{X}} = \mathbf{V}_{\ddot{\mathbf{x}}}^T \mathbf{\Lambda}_{\ddot{\mathbf{x}}} \mathbf{V}_{\ddot{\mathbf{x}}}$, we can construct $\boldsymbol{\varphi}_{\ddot{\mathbf{x}}} = [\varphi_1(\ddot{\mathbf{x}}), ..., \varphi_n(\ddot{\mathbf{x}})] \triangleq \mathbf{V}_{\ddot{\mathbf{x}}} \mathbf{\Lambda}_{\ddot{\mathbf{x}}}^{1/2}$, as an empirical kernel map for $\ddot{\mathbf{x}}$. Correspondingly, an *empirical* kernel map of the space $\mathcal{H}_{\ddot{\mathcal{X}}|\mathcal{Z}}$ is given by $\tilde{\boldsymbol{\varphi}}_{\ddot{\mathbf{x}}} = \mathbf{R}_Z \boldsymbol{\varphi}(\ddot{\mathbf{x}})$. Consequently, the centralized kernel matrix corresponding to the functions $\tilde{f}(\ddot{X})$ is

$$\widetilde{\mathbf{K}}_{\ddot{X}|Z} = \tilde{\boldsymbol{\varphi}}_{\ddot{\mathbf{x}}} \tilde{\boldsymbol{\varphi}}_{\ddot{\mathbf{x}}}^T = \mathbf{R}_Z \widetilde{\mathbf{K}}_{\ddot{X}} \mathbf{R}_Z. \quad (11)$$

Similarly, that corresponding to $\tilde{g}'$ is

$$\widetilde{\mathbf{K}}_{Y|Z} = \mathbf{R}_Z \widetilde{\mathbf{K}}_Y \mathbf{R}_Z. \quad (12)$$

Furthermore, we let the EVD decompositions of $\widetilde{\mathbf{K}}_{\ddot{X}|Z}$ and $\widetilde{\mathbf{K}}_{Y|Z}$ be $\widetilde{\mathbf{K}}_{\ddot{X}|Z} = \mathbf{E}_{\ddot{\mathbf{x}}|\mathbf{z}} \mathbf{\Lambda}_{\ddot{\mathbf{x}}|\mathbf{z}} \mathbf{E}_{\ddot{\mathbf{x}}|\mathbf{z}}^T$ and $\widetilde{\mathbf{K}}_{Y|Z} = \mathbf{E}_{\mathbf{y}|\mathbf{z}} \mathbf{\Lambda}_{\mathbf{y}|\mathbf{z}} \mathbf{E}_{\mathbf{y}|\mathbf{z}}^T$, respectively. $\mathbf{\Lambda}_{\ddot{\mathbf{x}}|\mathbf{z}}$ (resp. $\mathbf{\Lambda}_{\mathbf{y}|\mathbf{z}}$) is the diagonal matrix containing non-negative eigenvalues $\lambda_{\ddot{\mathbf{x}}|\mathbf{z},i}$ (resp. $\lambda_{\mathbf{y}|\mathbf{z},i}$). Let $\boldsymbol{\psi}_{\ddot{\mathbf{x}}|\mathbf{z}} = [\psi_{\ddot{\mathbf{x}}|\mathbf{z},1}(\ddot{\mathbf{x}}), ..., \psi_{\ddot{\mathbf{x}}|\mathbf{z},n}(\ddot{\mathbf{x}})] \triangleq \mathbf{V}_{\ddot{\mathbf{x}}|\mathbf{z}} \mathbf{\Lambda}_{\ddot{\mathbf{x}}|\mathbf{z}}^{1/2}$ and $\boldsymbol{\phi}_{\mathbf{y}|\mathbf{z}} = [\phi_{\mathbf{y}|\mathbf{z},1}(\ddot{\mathbf{y}}), ..., \phi_{\mathbf{y}|\mathbf{z},n}(\ddot{\mathbf{y}})] \triangleq \mathbf{V}_{\mathbf{y}|\mathbf{z}} \mathbf{\Lambda}_{\mathbf{y}|\mathbf{z}}^{1/2}$. We then have the following result which the proposed KCI-test is based on.

**Proposition 5 [Conditional independence test]**
*Under the null hypothesis $H_0$ (X and Y are conditionally independent given Z), we have that the statistic*

$$T_{CI} \triangleq \frac{1}{n} Tr(\widetilde{\mathbf{K}}_{\ddot{X}|Z} \widetilde{\mathbf{K}}_{Y|Z}) \quad (13)$$

*has the same asymptotic distribution as*

$$\check{T}_{CI} \triangleq \frac{1}{n} \sum_{k=1}^{n^2} \mathring{\lambda}_k \cdot z_k^2, \quad (14)$$

*where $\mathring{\lambda}_k$ are eigenvalues of $\check{\mathbf{w}}\check{\mathbf{w}}^T$ and $\check{\mathbf{w}} = [\check{\mathbf{w}}_1, ..., \check{\mathbf{w}}_n]$, with the vector $\check{\mathbf{w}}_t$ obtained by stacking $\check{\mathbf{M}}_t = [\psi_{\ddot{\mathbf{x}}|\mathbf{z},1}(\ddot{x}_t), ..., \psi_{\ddot{\mathbf{x}}|\mathbf{z},n}(\ddot{x}_t)]^T \cdot [\phi_{\mathbf{y}|\mathbf{z},1}(\ddot{y}_t), ..., \phi_{\mathbf{y}|\mathbf{z},n}(\ddot{y}_t)]$.* [4]

Similarly to the unconditional independence testing, we can do KCI-test by generating the approximate null distribution with Monte Carlo simulation. We first need to calculate $\widetilde{\mathbf{K}}_{\ddot{X}|Z}$ according to (11), $\widetilde{\mathbf{K}}_{Y|Z}$ according to (12), and their eigenvalues and eigenvectors. We then evaluate $T_{CI}$ according to (13) and simulate the distribution of $\check{T}_{CI}$ given by (14) by drawing

---
[4] Note that equivalently, $\mathring{\lambda}_k$ are eigenvalues of $\check{\mathbf{w}}^T\check{\mathbf{w}}$; hence there are at most $n$ non-zero values of $\mathring{\lambda}_k$.

i.i.d. $\chi_1^2$ samples and summing them up with weights $\mathring{\lambda}_k$. (For computational efficiency, in practice we drop all $\lambda_{\ddot{\mathbf{x}}|\mathbf{z},i}$, $\lambda_{\mathbf{y}|\mathbf{z},i}$, and $\mathring{\lambda}_k$ which are smaller than $10^{-5}$). Finally the p-value is calculated as the probability of $\check{T}_{CI}$ exceeding $T_{CI}$. Approximating the null distribution with a Gamma distribution, which is given next, avoids calculating $\mathring{\lambda}_k$ and simulating the null distribution, and is computationally more efficient.

### 3.4 Approximating the null distribution by a Gamma distribution

In addition to the simulation-based method to find the null distribution, as in Gretton et al. (2008), we provide approximations to the null distributions with a two-parameter Gamma distribution. The two parameters of the Gamma distribution are related to the mean and variance. In particular, under the null hypothesis that $X$ and $Y$ are independent (resp. conditionally independent given $Z$), the distribution of $\check{T}_{UI}$ given by (9) (resp. of $\check{T}_{CI}$ given by (14)) can be approximated by the $\Gamma(k, \theta)$ distribution:

$$p(t) = t^{k-1} \frac{e^{-t/\theta}}{\theta^k \Gamma(k)},$$

where $k = \mathbb{E}^2(\check{T}_{UI})/\mathbb{V}ar(\check{T}_{UI})$ and $\theta = \mathbb{V}ar(\check{T}_{UI})/\mathbb{E}(\check{T}_{UI})$ in the unconditional case, and $k = \mathbb{E}^2(\check{T}_{CI})/\mathbb{V}ar(\check{T}_{CI})$ and $\theta = \mathbb{V}ar(\check{T}_{CI})/\mathbb{E}(\check{T}_{CI})$ in the conditional case. The means and variances of $\check{T}_{UI}$ and $\check{T}_{CI}$ on the given sample $\mathcal{D}$ are given in the following proposition.

**Proposition 6** *i. Under the null hypothesis that $X$ and $Y$ are independent, on the given sample $\mathcal{D}$, we have*

$$\mathbb{E}(\check{T}_{UI}|\mathcal{D}) = \frac{1}{n^2} Tr(\widetilde{\mathbf{K}}_X) \cdot Tr(\widetilde{\mathbf{K}}_Y), \text{ and}$$

$$\mathbb{V}ar(\check{T}_{UI}|\mathcal{D}) = \frac{2}{n^4} Tr(\widetilde{\mathbf{K}}_X^2) \cdot Tr(\widetilde{\mathbf{K}}_Y^2).$$

*ii. Under the null hypothesis of $X \perp\!\!\!\perp Y|Z$, we have*

$$\mathbb{E}\{\check{T}_{CI}|\mathcal{D}\} = \frac{1}{n} Tr(\check{\mathbf{w}}\check{\mathbf{w}}^T), \text{ and}$$

$$\mathbb{V}ar(\check{T}_{IC}|\mathcal{D}) = \frac{2}{n^2} Tr[(\check{\mathbf{w}}\check{\mathbf{w}}^T)^2].$$

### 3.5 Practical issues: On determination of the hyperparameters

In unconditional independence testing, the hyperparameters are the kernel widths for constructing the kernel matrices $\widetilde{\mathbf{K}}_X$ and $\widetilde{\mathbf{K}}_Y$. We found that the performance is robust to these parameters within a certain range; in particular, as in Gretton et al. (2008), we use the median of the pairwise distances of the points.

In KCI-test, the hyperparameters fall into three categories. The first one includes the kernel widths used to construct the kernel matrices $\widetilde{\mathbf{K}}_{\ddot{X}}$ and $\widetilde{\mathbf{K}}_Y$, which are used later to form the matrices $\widetilde{\mathbf{K}}_{\ddot{X}|Z}$ and $\widetilde{\mathbf{K}}_{Y|Z}$ according to (11) and (12). The values of the kernel widths should be such that one can use the information in the data effectively. In our experiments we normalize all variables to unit variance, and use some empirical values for those kernel widths: they are set to 0.8 if the sample size $n \leq 200$, to 0.3 if $n > 1200$, or to 0.5 otherwise. We found that this simple setting always works well in all our experiments.

The other two categories contain the kernel widths for constructing $\widetilde{\mathbf{K}}_Z$ and the regularization parameter $\varepsilon$, which are needed in calculating $\mathbf{R}_Z$ in (10). These parameters should be selected carefully, especially when the conditioning set $Z$ is large. If they are too large, the corresponding regression functions $h_f^*$ and $h_{g'}^*$ may underfit, resulting in a large probability of Type I errors (where the CI hypothesis is incorrectly rejected). On the contrary, if they are too small, these regression function may overfit, which may increase the probability of Type II errors (where the CI hypothesis is not rejected although being false). Moreover, if $\varepsilon$ is too small, $\mathbf{R}_Z$ in (10) tends to vanish, resulting in very small values in $\widetilde{\mathbf{K}}_{\ddot{X}|Z}$ and $\widetilde{\mathbf{K}}_{Y|Z}$, and then the performance may be deteoriated by rounding errors.

We found that when the dimensionality of $Z$ is small (say, one or two variables), the proposed method works well even with some simple empirical settings (say, $\varepsilon = 10^{-3}$, and the kernel width for constructing $\widetilde{\mathbf{K}}_Z$ equals half of that for constructing $\widetilde{\mathbf{K}}_{\ddot{X}}$ and $\widetilde{\mathbf{K}}_Y$). When $Z$ contains many variables, to make the regression functions $h_f^*$ and $h_{g'}^*$ more flexible, such that they could properly capture the information about $Z$ in $f(\ddot{X}) \in \mathcal{H}_{\ddot{X}}$ and $g'(Y) \in \mathcal{H}_Y$, respectively, we use separate regularization parameters and kernel widths for them, denoted by $\{\varepsilon_f, \boldsymbol{\sigma}_f\}$ and $\{\varepsilon_{g'}, \boldsymbol{\sigma}_{g'}\}$. To avoid both overfitting and underfitting, we extend the Gaussian process (GP) regression framework to the multi-output case, and learn these hyperparameters by maximizing the total marginal likelihood. Details are skipped. The MATLAB source code is available at http://people.tuebingen.mpg.de/kzhang/KCI-test.zip.

## 4 Experiments

We apply the proposed method KCI-test to both synthetic and real data to evaluate its practical performance and compare it with $\text{CI}_{\text{PERM}}$ (Fukumizu et al., 2008). It is also used for causal discovery.

### 4.1 On the effect of the dimensionality of $Z$ and the sample size

We examine how the probabilities of Type I and II errors of KCI-test change along with the size of the conditioning set $Z$ ($D = 1, 2, ..., 5$) and the sample size ($n = 200$ and $400$) *in particular situations* with simulations. We consider two cases as follows.

In Case I, only one variable in $Z$, namely, $Z_1$, is effective, i.e., other conditioning variables are independent from $X$, $Y$, and $Z_1$. To see how well the derived asymptotic null distribution approximates the true one, we examined if the probability of Type I errors is consistent with the significance level $\alpha$ that is specified in advance. We generated $X$ and $Y$ from $Z_1$ accroding to the post-nonlinear data generating procedure (Zhang and Hyvärinen, 2009): they were constructed as $G(F(Z_1) + E)$ where $G$ and $F$ are random mixtures of linear, cubic, and tanh functions and are different for $X$ and $Y$, and $E$ is independent across $X$ and $Y$ and has various distributions. Hence $X \perp\!\!\!\perp Y|Z$ holds. In our simulations $Z_i$ were i.i.d. Gaussian.

A good test is expected to have small probability of Type II errors. To see how large it is for KCI-test, we also generated the data which do not follow $X \perp\!\!\!\perp Y|Z$ by adding the same variable to $X$ and $Y$ produced above. We increased the dimensionality of $Z$ and the sample size $n$, and repeated the CI tests 1000 random replications. Figure 1 (a,b) plots the resulting probability of Type I errors and that of Type II errors at the significance level $\alpha = 0.01$ and $\alpha = 0.05$, respectively.

In Case II, all variables in the conditioning set $Z$ are effective in generating $X$ and $Y$. We first generated the independent variables $Z_i$, and then similarly to Case I, to examine Type I errors, $X$ and $Y$ were generated as $G(\sum_i F_i(Z_i) + E)$. To examine Type II errors, we further added the same variable to $X$ and $Y$ such that they became conditionally dependent given $Z$. Figure 1 (c,d) gives the probabilities of Type I and Type II errors of KCI-test obtained on 1000 random replications.

One can see that with the derived null distribution, the 1% and 5% quantiles are approximated very well for both sample sizes, since the resulting probabilities of Type I errors are very close to the significance levels. Gamma approximation tends to produce slightly larger probabilities of Type I errors, meaning that the two-parameter Gamma approximation may have a slightly lighter tail than the true null distribution. With a fixed significance level $\alpha$, as $D$ increases, the probability of Type II errors always increases. This is intuitively reasonable: due to the finite sample size effect, as the conditioning set becomes larger and larger, $X$ and $Y$ tend to be considered as conditionally inde-pendent. On the other hand, as the sample size increases from 200 to 400, the probability of Type II errors quickly approaches zero.

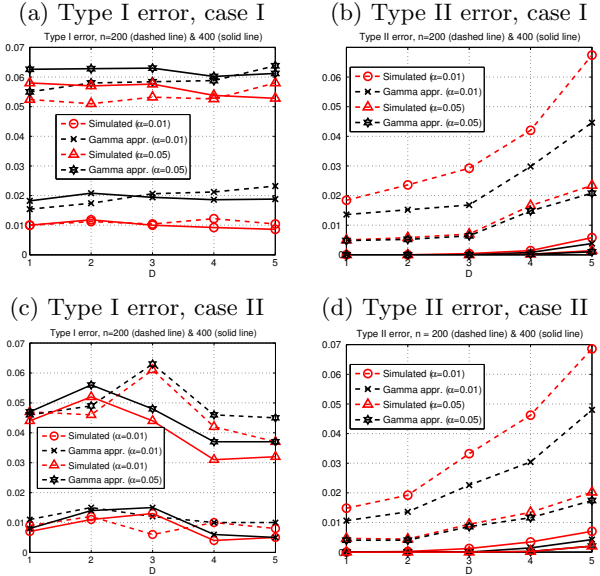

(a) Type I error, case I  (b) Type II error, case I

(c) Type I error, case II  (d) Type II error, case II

Figure 1: The probability of Type I and Type II errors obtained by simulations in various situations (dashed line: $n = 200$; solid line: $n = 400$). Top: Case I (only one variable in $Z$ is effective to $X$ and $Y$). Bottom: Case II (all variables in $Z$ are effective). Note that for a good testing method, the propobability of Type I errors is close to the significance level $\alpha$, and that of Type II errors is as small as possible.

We compared KCI-test with $\text{CI}_{\text{PERM}}$ (with the standard setting of 500 bootstrap samples) in terms of both types of errors and computational efficiency. For conciseness, we only report the probabilities of Type I and II errors of $\text{CI}_{\text{PERM}}$ in Case II; see Figure 2 (a). One can see even when $D = 1$, the probability of Type I errors is clearly larger than the corresponding significance level. Furthermore, it is very sensitive to $D$ and $n$. As $D$ becomes large, say, greater than 3, The probabilities of Type II errors increase rapidly to 1, i.e., the test almost always fails to reject the CI hypothesis when it is actually false. It seems that the sample size is too small for $\text{CI}_{\text{PERM}}$ to give reliable results. Figure 2 (b) shows the average CPU time taken by KCI-test and $\text{CI}_{\text{PERM}}$ (note that it is in log scale). KCI-test is computationally more efficient, especially when $D$ is large. Because we use the GP regression framework to learn hyperparameters, which KCI-test spends most of the time on, the computational load of KCI-test is more sensitive to $n$.[5] On the other hand,

---

[5] As claimed in Sec. 3.5, when $D$ is not high, say, $D = 1$ or 2, even fixed empirical values of the hyperparameters work very well. However, for consistentcy of the compari-

it is far less sensitive to $D$.[6]

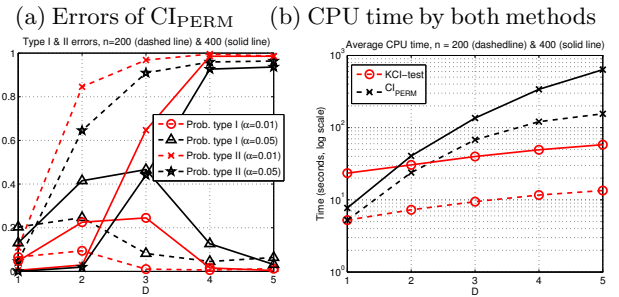

(a) Errors of $\text{CI}_{\text{PERM}}$   (b) CPU time by both methods

Figure 2: Results of $\text{CI}_{\text{PERM}}$ shown for comparison. (a) Probabilities of both Type I and II errors in Case II (dashed line: $n = 200$; solid line: $n = 400$). (b) Average CPU time taken by KCI-test and $\text{CI}_{\text{PERM}}$ in log scale. Note that for KCI-test, we include both the time for simulating the null distribution and that for Gamma approximation.

### 4.2 Application in causal discovery

CI tests are frequently used in problems of causal inference: in those problems, one assumes that the true causal structure of $n$ random variables $X_1, \ldots, X_n$ can be represented by a directed acyclic graph (DAG) $\mathcal{G}$. More specifically, the *causal Markov condition* assumes that the joint distribution satisfies all CIs that are imposed by the true causal graph (note that this is an assumption about the physical generating process of the data, not only about their distribution). So-called constraint-based methods like the PC algorithm (Spirtes et al., 2001) make the additional assumption of faithfullness (i.e., the joint distribution does not allow any CIs that are not entailed by the Markov condition) and recover the graph structure by exploiting the (conditional) independences that can be found in the data. Obviously, this is only possible up to *Markov equivalence classes*, which are sets of graphs that impose exactly the same independences and CIs. It is well known that small mistakes at the beginning of the algorithm (e.g. missing an independence relation) may lead to significant errors in the resulting DAG. Therefore the performance of those methods relies heavily on (conditional) independence testing methods.

For continuous data, the PC algorithm can be applied

---

son, we always used GP for hyperparameter learning.

[6] To see the price our test pays for generality, in another simulation, we considered the linear Gaussian case and compared KCI-test with the partial correlated based one. We found that for the particular problems we investigated, both methods give very similar type I errors; partial correlation gives much smaller type II errors than KCI-test, which is natural since KCI-test applies to far more general situations. Details are skipped due to space limitation.

using partial correlation or mutual information. The former assumes linear relationships and Gaussian distributions, and the latter does not lead to a significance test. Sun et al. (2007) and Tillman et al. (2009) propose to use $\text{CI}_{\text{PERM}}$ for CI testing in PC. Based on the promising results of KCI-test, we propose to also apply it to causal inference.

### 4.2.1 Simulated data

We generated data from a random DAG $\mathcal{G}$. In particular, we randomly chose whether an edge exists and sampled the functions that relate the variables from a Gaussian Process prior. We sampled four random variables $X_1, \ldots, X_4$ and allowed arrows from $X_i$ to $X_j$ only for $i < j$. With probability 0.5 each possible arrow is either present or absent. If arrows exist, from $X_1$ and $X_3$ to $X_4$, say, we sample $X_4$ from a Gaussian Process with mean function $U_1 \cdot X_1 + U_3 \cdot X_3$ (with $U_1, U_3 \overset{\text{iid}}{\sim} U[-2;2]$) and a Gaussian kernel (with each dimension randomly weighted between 0.1 and 0.6) plus a noise kernel as the covariance function. For significance level 0.01 and sample sizes between 100 and 700 we simulated 100 DAGs, and checked how often the different methods infer the correct Markov equivalence class. Figure 3 shows how often PC based on KCI-test, $\text{CI}_{\text{PERM}}$, or partial correlation recovered the correct Markov equivalence class. PC based on KCI-test gives clearly the best results.

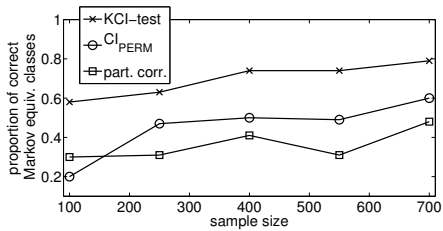

Figure 3: The chance that the correct Markov equivalence class was inferred with PC combined with different CI testing methods. KCI-test outperforms $\text{CI}_{\text{PERM}}$ and partial correlation.

### 4.2.2 Real data

We applied our method to continuous variables in the Boston Housing data set, which is available at the UCI Repository (Asuncion and Newman, 2007). Due to the large number of variables we choose the significance level to be 0.001, as a rough way to correct for multiple testing. Figure 4 shows the results for PC using $\text{CI}_{\text{PERM}}$ ($\text{PC}_{\text{CI}_{\text{PERM}}}$) and KCI-test ($\text{PC}_{\text{KCI-test}}$). For conciseness, we report them in the same figure: the *red* arrows are the ones inferred by $\text{PC}_{\text{CI}_{\text{PERM}}}$ and all *solid* lines show the result by $\text{PC}_{\text{KCI-test}}$. Ergo, red solid lines were found by both methods. Please refer to the data set for the explanation of the variables. Although one can argue about the ground truth for this data set, we regard it as promising that our method finds links between number of rooms (RM) and median value of houses (MED) and between non-retail business (IND) and nitric oxides concentration (NOX). The latter is also missing in the result on these data given by Margaritis (2005); instead their method gives some dubious links like crime rate (CRI) to nitric oxides (NOX), for example.

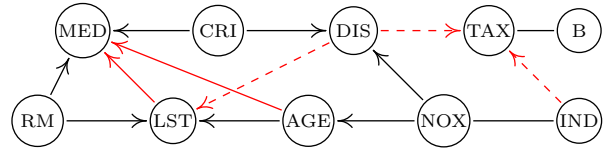

Figure 4: Outcome of the PC algorithm applied to the continuous variables of the Boston Housing Data Set (red lines: $\text{PC}_{\text{CI}_{\text{PERM}}}$, solid lines: $\text{PC}_{\text{KCI-test}}$).

## 5  Conclusion

We proposed a novel method for conditional independence testing. It makes use of the characterization of conditional independence in terms of uncorrelatedness of functions in suitable reproducing kernel Hilbert spaces, and the proposed test statistic can be easily calculated from the kernel matrices. We derived its distribution under the null hypothesis of conditional independence. This distribution can either be generated by Monte Carlo simulation or approximated by a two-parameter Gamma distribution. Compared to discretization-based conditional independence testing methods, the proposed one exploits more complete information of the data, and compared to those methods defining the test statistic in terms of the estimated conditional densities, calculation of the proposed statistic involves less random errors. We applied the method on both simulated and real world data, and the results suggested that the method outperforms existing techniques in both accuracy and speed.

#### Acknowledgement

ZK thanks Stefanie Jegelka for helpful discussions. DJ was supported by DFG (SPP 1395).

## Appendix: Proofs

The following lemmas will be used in the proof of Theorem 3. Consider the eigenvalues $\lambda_{X,i}$ and eigenvectors $\mathbf{V}_{\mathbf{x},i}$ of the kernel matrix $\widetilde{\mathbf{K}}_X$ and the eigenvalues $\lambda^*_{X,i}$ and normalized eigenfunctions $u_{X,i}$ (which have unit variance) of the corresponding kernel $k_{\mathcal{X}}$. Let $\check{\mathbf{x}}$ be a fixed-size subsample of $\mathbf{x}$. The following lemma is due to Theorems 3.4 and 3.5 of Baker (1977).

**Lemma 7** *i. $\frac{1}{n}\lambda_{X,i}$ converge in probability to $\lambda_{X,i}^*$.*

*ii. For the simple eigenvalue $\lambda_{X,i}^*$, whose algebraic multiplicity is 1, $\sqrt{n}\mathbf{V}_{\mathbf{x},i}(\check{\mathbf{x}})$ converges in probability to $u_{X,i}(\check{\mathbf{x}})$, where $\mathbf{V}_{\mathbf{x},i}(\check{\mathbf{x}})$ denotes the values of $\mathbf{V}_{\mathbf{x},i}$ corresponding to the subsample $\check{\mathbf{x}}$.*

Suppose that the eigenvalues $\lambda_{X,k}^*, \lambda_{X,k+1}^*, ..., \lambda_{X,k+r}^*$ ($r \geq 1$) are the same and different from any other eigenvalue. The corresponding eigenfunctions are then non-unique. Denote by $\vec{u}_{X,k:k+r} \triangleq (u_{X,k}, u_{X,k+1}, ..., u_{X,k+r})^T$ an arbitrary vector of the eigenfunctions corresponding to this eigenvalue with multiplicity $r+1$. Let $\mathcal{S}_{X,k}$ be the space of $\vec{u}_{X,k:k+r}$.

**Lemma 8** *i. The distance from $\sqrt{n}\mathbf{V}_{\mathbf{x},k+q}(\check{\mathbf{x}})$ ($0 \leq q \leq r$) to the corresponding points in the space $\mathcal{S}_{X,k}$ converges to zero in the following sense:*

$$\inf\{||\sqrt{n}\mathbf{V}_{\mathbf{x},k+q}(\check{\mathbf{x}}) - u'(\check{\mathbf{x}})|| \, | \, u'(x) \in \mathcal{S}_{X,k}\} \to 0, \text{ as } n \to \infty.$$

*ii. There exists an $(r+1) \times (r+1)$ orthogonal matrix $\mathbf{P}_n$, which may depend on $n$, such that $\sqrt{n} \cdot \mathbf{P}_n \cdot [\mathbf{V}_{\mathbf{x},k}(\check{\mathbf{x}}), \mathbf{V}_{\mathbf{x},k+1}(\check{\mathbf{x}}), ..., \mathbf{V}_{\mathbf{x},k+r}(\check{\mathbf{x}})]^T$ converges in probability to $\vec{u}_{X,k:k+r}(\check{\mathbf{x}})$.*

Item *(i.)* in the above lemma is a reformulation of Theorem 3.6 of Baker (1977), while item *(ii.)* is its straightforward consequence.

The following lemma is a reformulation of Theorem 4.2 of Billingsley (1999).

**Lemma 9** *Let $\{A_{L,n}\}$ be a double sequence of random variables with indices $L$ and $n$, $\{B_L\}$ and $\{C_n\}$ sequences of random variables, and $D$ a random variable. Assume that they are defined in a separable probability space. Suppose that, for each $L$, $A_{L,n} \xrightarrow{d} B_L$ as $n \to \infty$ and that $B_L \xrightarrow{d} D$ as $L \to \infty$. Suppose further that $\lim \limsup_{L \to \infty, n \to \infty} P(\{|A_{L,n} - C_n| \geq \varepsilon\}) = 0$ for each positive $\varepsilon$. Then $C_n \xrightarrow{d} D$ as $n \to \infty$.*

**Sketch of proof of Theorem 3.** We first define a $L \times L$ block-diagonal matrix $\mathbf{P}_X$ as follows. For simple eigenvalues $\lambda_{X,i}^*$, $\mathbf{P}_{X,ii} = 1$, and all other entries in the $i$th row and column are zero. For the eigenvalues with multiplicity $r+1$, say, $\lambda_{X,k}^*, \lambda_{X,k+1}^*, ..., \lambda_{X,k+r}^*$, the corresponding main diagonal block from the $k$th row to the $(k+r)$th row of $\mathbf{P}_X$ is an orthogonal matrix. According to Lemmas 7 and 8 and the continuous mapping theorem (CMT) (Mann and Wald, 1943), which states that continuous functions are limit-preserving even if their arguments are sequences of random variables, there exists $\mathbf{P}_X$, which may depend on $n$, such that $\mathbf{P}_X \cdot [\psi_1(x_t), ..., \psi_L(x_t)]^T \to [\sqrt{\lambda_{X,1}^*}u_{X,1}(x_t), ..., \sqrt{\lambda_{X,L}^*}u_{X,L}(x_t)]^T$ in probability as $n \to \infty$. Similary, we can define the orthogonal matrix $\mathbf{P}_Y$ such that $\mathbf{P}_Y \cdot [\phi_1(y_t), ..., \phi_L(y_t)]^T \to [\sqrt{\lambda_{Y,1}^*}u_{Y,1}(y_t), ..., \sqrt{\lambda_{Y,L}^*}u_{Y,L}(y_t)]^T$ in probability as $n \to \infty$.

Let $\mathbf{v}_t$ be the random vector obtained by stacking the random $L \times L$ matrix $\mathbf{P}_X \cdot \mathbf{M}_t \cdot \mathbf{P}_Y^T$ with $M_{ij,t} = \psi_i(x_t) \cdot \phi_j(y_t)$ as the $(i,j)$th entry of $\mathbf{M}_t$. One can see that

$$\frac{1}{n}\sum_{t=1}^n ||\mathbf{v}_t||^2 = \frac{1}{n}\sum_{t=1}^n \text{Tr}(\mathbf{P}_X \mathbf{M}_t \mathbf{P}_Y^T \cdot \mathbf{P}_Y \mathbf{M}_t^T \mathbf{P}_X^T)$$
$$= \frac{1}{n}\sum_{t=1}^n \text{Tr}(\mathbf{M}_t \mathbf{M}_t^T) = \sum_{i,j=1}^L S_{ij}^2. \quad (15)$$

Again, according to CMT, one can see that $\mathbf{v}_t$ converges in probability to $\mathbf{w}_t$. Furthermore, according to the vector-valued CLT (Eicker, 1966), as $x_t$ and $y_t$ are i.i.d., the vector $\frac{1}{\sqrt{n}}\sum_{t=1}^n \mathbf{w}_t$ converges in distribution to a multivariate normal distribution as $n \to \infty$. Because of CMT, $\frac{1}{\sqrt{n}}\sum_{t=1}^n \mathbf{v}_t$ then converges to the same normal distribution as $n \to \infty$. As $f(X) \in \mathcal{H}_{\mathcal{X}}$ and $g(Y) \in \mathcal{H}_{\mathcal{Y}}$ are uncorrelated, we know that $u_{X,i}$ and $u_{Y,j}$ are uncorrelated, and consequently the mean of this normal distribution is $\mathbb{E}(\mathbf{w}_t) = \mathbf{0}$. The covariance is $\mathbf{\Sigma} = \mathbb{C}\text{ov}(\frac{1}{\sqrt{n}}\sum_{t=1}^n \mathbf{w}_t) = \mathbb{C}\text{ov}(\mathbf{w}_t) = \mathbb{E}(\mathbf{w}_t\mathbf{w}_t^T)$.

Assume that we have the EVD decompostion $\mathbf{\Sigma} = \mathbf{V}_{\mathbf{w}}\mathbf{\Lambda}_{\mathbf{w}}\mathbf{V}_{\mathbf{w}}^T$, where $\mathbf{\Lambda}_{\mathbf{w}}$ is the diagonal matrix containing non-negative eigenvalues $\mathring{\lambda}_k^*$. Let $\mathbf{v}'_t \triangleq \mathbf{V}_{\mathbf{w}}^T\mathbf{v}_t$. Clearly $\frac{1}{\sqrt{n}}\sum_{t=1}^n \mathbf{v}'_t$ follows $\mathcal{N}(\mathbf{0}, \mathbf{\Lambda}_{\mathbf{w}})$ asympotically. That is,

$$\frac{1}{n}\sum_{t=1}^n ||\mathbf{v}_t||^2 = \frac{1}{n}\sum_{t=1}^n ||\mathbf{v}'_t||^2 \xrightarrow{d} \sum_{k=1}^{L^2} \mathring{\lambda}_k^* z_k^2. \quad (16)$$

Combining (15) and (16) gives (5).

If $X$ and $Y$ are independent, for $k \neq i$ or $l \neq j$, one can see that the non-diagonal entries of $\mathbf{\Sigma}$ are $\mathbb{E}[\sqrt{\lambda_{X,i}^*\lambda_{Y,j}^*\lambda_{X,k}^*\lambda_{Y,l}^*}u_{X,i}(x_t)u_{Y,j}(y_t)u_{X,k}(x_t)u_{Y,l}(y_t)] = \sqrt{\lambda_{X,i}^*\lambda_{Y,j}^*\lambda_{X,k}^*\lambda_{Y,l}^*}\mathbb{E}[u_{X,i}(x_t)u_{X,k}(x_t)]\mathbb{E}[u_{Y,j}(y_t)u_{Y,l}(y_t)] = 0$. The diagonal entries of $\mathbf{\Sigma}$ are $\lambda_{X,i}^*\lambda_{Y,j}^* \cdot \mathbb{E}[u_{X,i}^2(x_t)]\mathbb{E}[u_{Y,j}^2(y_t)] = \lambda_{X,i}^*\lambda_{Y,j}^*$, which are also eigenvalues of $\mathbf{\Sigma}$. Substituting this result into (5), one obtains (6).

Finally, consider Lemma 9, and let $A_{L,n} \triangleq \sum_{i,j=1}^L S_{ij}^2$, $B_L \triangleq \sum_{k=1}^{L^2} \mathring{\lambda}_k^* z_k^2$, $C_n \triangleq \sum_{i,j=1}^n S_{ij}^2$, and $D \triangleq \sum_{k=1}^{\infty} \mathring{\lambda}_k^* z_k^2$. One can then see that $\sum_{i,j=1}^n S_{ij}^2 \xrightarrow{d}$

$\sum_{k=1}^{\infty} \mathring{\lambda}_k^* z_k^2$ as $n \to \infty$. That is, (5) also holds as $L = n \to \infty$. As a special case of (5), (6) also holds as $L = n \to \infty$. ∎

**Sketch of proof of Theorem 4.** On the one hand, due to (7), we have $T_{UI} = \sum_{i,j=1}^{n} S_{ij}^2$, which, according to (6), converges in distribution to $\sum_{i,j=1}^{\infty} \lambda_{X,i}^* \lambda_{Y,j}^* \cdot z_{ij}^2$ as $n \to \infty$. On the other hand, by extending the proof of Theorem 1 of Gretton et al. (2009), one can show that $\sum_{i,j=1}^{\infty} (\frac{1}{n^2} \lambda_{\mathbf{x},i} \lambda_{\mathbf{y},j} - \lambda_{X,i}^* \lambda_{Y,j}^*) z_{ij}^2 \to 0$ in probability as $n \to \infty$. That is, $\check{T}_{UI}$ converges in probability to $\sum_{i,j=1}^{\infty} \lambda_{X,i}^* \lambda_{Y,j}^* z_{ij}^2$ as $n \to \infty$. Consequently, $T_{UI}$ and $\check{T}_{UI}$ have the same asymptotic distribution. ∎

**Sketch of proof of Proposition 5.** Here we let $X$, $Y$, $\widetilde{\mathbf{K}}_X$, and $\widetilde{\mathbf{K}}_Y$ in Theorem 3 be $\ddot{X}$, $(Y, Z)$, $\widetilde{\mathbf{K}}_{\ddot{X}|Z}$, and $\widetilde{\mathbf{K}}_{Y|Z}$, respectively. According to (7) and Theorem 3, one can see that $T_{CI} \xrightarrow{d} \sum_{k=1}^{\infty} \mathring{\lambda}_k^* z_k^2$ as $n \to \infty$. Again, we extend the proof of Theorem 1 of Gretton et al. (2009) to show that $(\sum_{k=1}^{\infty} \frac{1}{n} \mathring{\lambda}_k - \mathring{\lambda}_k^*) z_k^2 \to 0$ as $n \to \infty$, or that $\sum_{k=1}^{\infty} \frac{1}{n} \mathring{\lambda}_k z_k^2 \to \sum_{k=1}^{\infty} \mathring{\lambda}_k^* z_k^2$ in probability as $n \to \infty$. The key step is to show that $\sum_k |\frac{1}{n} \mathring{\lambda}_k - \mathring{\lambda}_k^*| \to 0$ in probability as $n \to \infty$. Details are skipped. ∎

**Sketch of proof of Proposition 6.** As $z_{ij}^2$ follow the $\chi^2$ distribution with one degree of freedom, we have $\mathbb{E}(z_{ij}^2) = 1$ and $\mathbb{V}ar(z_{ij}^2) = 2$. According to (9), we have $\mathbb{E}(\check{T}_{UI}|\mathcal{D}) = \frac{1}{n^2} \sum_{i,j} \lambda_{\mathbf{x},i} \lambda_{\mathbf{y},j} = \frac{1}{n^2} \sum_i \lambda_{\mathbf{x},i} \sum_j \lambda_{\mathbf{y},j} = \frac{1}{n^2} \text{Tr}(\widetilde{\mathbf{K}}_X) \cdot \text{Tr}(\widetilde{\mathbf{K}}_Y)$. Furthermore, bearing in mind that $z_{ij}^2$ are independent variables across $i$ and $j$, and recalling $\text{Tr}(\mathbf{K}_X^2) = \sum_i \lambda_{\mathbf{x},i}^2$, one can see that $\mathbb{V}ar(\check{T}_{UI}|\mathcal{D}) = \frac{1}{n^4} \sum_{i,j} \lambda_{\mathbf{x},i}^2 \lambda_{\mathbf{y},j}^2 \mathbb{V}ar(z_{ij}^2) = \frac{2}{n^4} \sum_i \lambda_{\mathbf{x},i}^2 \sum_j \lambda_{\mathbf{y},j}^2 = \frac{2}{n^4} \text{Tr}(\widetilde{\mathbf{K}}_X^2) \cdot \text{Tr}(\widetilde{\mathbf{K}}_Y^2)$. Consequently $(i)$ is true.

Similarly, from (14), one can calculate the mean $\mathbb{E}\{\check{T}_{CI}|\mathcal{D}\}$ and variance $\mathbb{V}ar(\check{T}_{IC}|\mathcal{D})$, as given in $(ii)$. ∎

# References

A. Asuncion and D.J. Newman. UCI machine learning repository. http://archive.ics.uci.edu/ml/, 2007.

C. Baker. *The numerical treatment of integral equations.* Oxford University Press, 1977.

W. P. Bergsma. *Testing conditional independence for continuous random variables*, 2004. EURANDOM-report 2004-049.

P. Billingsley. *Convergence of Probability Measures.* John Wiley and Sons, 1999.

J. J. Daudin. Partial association measures and an application to qualitative regression. *Biometrika*, 67:581–590, 1980.

A. P. Dawid. Conditional independence in statistical theory. *Journal of the Royal Statistical Society. Series B*, 41:1–31, 1979.

F. Eicker. A multivariate central limit theorem for random linear vector forms. *The Annals of Mathematical Statistics*, 37:1825–1828, 1966.

K. Fukumizu, F. R. Bach, M. I. Jordan, and C. Williams. Dimensionality reduction for supervised learning with reproducing kernel hilbert spaces. *JMLR*, 5:2004, 2004.

K. Fukumizu, A. Gretton, X. Sun, and B. Schölkopf. Kernel measures of conditional dependence. In *NIPS 20*, pages 489–496, Cambridge, MA, 2008. MIT Press.

A. Gretton, K. Fukumizu, C. H. Teo, L. Song, B. Schölkopf, and A. J. Smola. A kernel statistical test of independence. In *NIPS 20*, pages 585–592, Cambridge, MA, 2008.

A. Gretton, K. Fukumizu, Z. Harchaoui, and B. K. Sriperumbudur. A fast, consistent kernel two-sample test. In *NIPS 23*, pages 673–681, Cambridge, MA, 2009. MIT Press.

T. M. Huang. Testing conditional independence using maximal nonlinear conditional correlation. *Ann. Statist.*, 38:2047–2091, 2010.

D. Koller and N. Friedman. *Probabilistic Graphical Models: Principles and Techniques.* MIT Press, Cambridge, MA, 2009.

A. J. Lawrance. On conditional and partial correlation. *The American Statistician*, 30:146–149, 1976.

O. Linton and P. Gozalo. *Conditional independence restrictions: testing and estimation*, 1997. Cowles Foundation Discussion Paper 1140.

H. B. Mann and A. Wald. On stochastic limit and order relationships. *The Annals of Mathematical Statistics*, 14:217–226, 1943.

D. Margaritis. Distribution-free learning of bayesian network structure in continuous domains. In *Proc. AAAI 2005*, pages 825–830, Pittsburgh, PA, 2005.

J. Pearl. *Causality: Models, Reasoning, and Inference.* Cambridge University Press, Cambridge, 2000.

B. Schölkopf and A. Smola. *Learning with kernels.* MIT Press, Cambridge, MA, 2002.

K. Song. Testing conditional independence via rosenblatt transforms. *Ann. Statist.*, 37:4011–4045, 2009.

P. Spirtes, C. Glymour, and R. Scheines. *Causation, Prediction, and Search.* MIT Press, Cambridge, MA, 2nd edition, 2001.

L. Su and H. White. A consistent characteristic function-based test for conditional independence. *Journal of Econometrics*, 141:807–834, 2007.

L. Su and H. White. A nonparametric hellinger metric test for conditional independence. *Econometric Theory*, 24:829–864, 2008.

X. Sun, D. Janzing, B. Schölkopf, and K. Fukumizu. A kernel-based causal learning algorithm. In *Proc. ICML 2007*, pages 855–862. Omnipress, 2007.

R. Tillman, A. Gretton, and P. Spirtes. Nonlinear directed acyclic structure learning with weakly additive noise models. In *NIPS 22*, Vancouver, Canada, 2009.

K. Zhang and A. Hyvärinen. On the identifiability of the post-nonlinear causal model. In *Proc. UAI 25*, Montreal, Canada, 2009.